\documentclass[a4paper]{article}

\usepackage[utf8]{inputenc}
\usepackage{INTERSPEECH_v2}
\usepackage{multirow}
\usepackage{hyperref}
\usepackage{color}
\usepackage{comment}
\usepackage{mathtools}
\usepackage{booktabs}
\usepackage{tikz}%Pour les plots avec latex
\usetikzlibrary{patterns} %Pour les hachures
\usepackage{tipa}
\usepackage[export]{adjustbox}

\definecolor{codegreen}{rgb}{0,0.7,0}
\definecolor{codegray}{rgb}{0.5,0.5,0.5}
\definecolor{codepurple}{rgb}{0.58,0,0.82}
\definecolor{backcolour}{rgb}{0.95,0.95,0.92}

\title{Learning weakly supervised multimodal phoneme embeddings}
\name{Rahma Chaabouni$^1$, Ewan Dunbar$^1$, Neil Zeghidour$^{1,2}$,Emmanuel Dupoux$^1$}
\address{
  $^1$Département d'Etudes Cognitives, Ecole Normale Supérieure, Ecole des Hautes Etudes en Sciences Sociales, PSL Research University, Centre National de la Recherche Scientifique, Paris, France.
  $^2$Facebook AI Research, Paris}
\email{chaabounirahma,emmanuel.dupoux, neil.zeghidour@gmail.com,emd@umd.edu}

\begin{document}

\maketitle
% ED: salut, j'aurais tendance à partir dans l'autre sens: dire qu'il ya  
% plusieurs papiers qui ont fait du multimodal, mais qu'aucun ne l'a 
% fait en weakly supervised. Mais, bon, ça ne change pas grand chose
%190 words
\begin{abstract}
Recent works have explored deep architectures for learning multimodal speech representation (e.g. audio and images, articulation and audio) in a supervised way. Here we investigate the role of combining different speech modalities, i.e. audio and visual information representing the lips’ movements, in a weakly supervised way using Siamese networks and lexical same-different side information. In particular, we ask whether one modality can benefit from the other to provide a richer representation for phone recognition in a weakly supervised setting. We introduce mono-task and multi-task methods for merging speech and visual modalities for phone recognition. The mono-task learning consists in applying a Siamese network on the concatenation of the two modalities, while the multi-task learning receives several different combinations of modalities at train time. We show that multi-task learning enhances discriminability for visual and multimodal inputs while minimally impacting auditory inputs. Furthermore, we present a qualitative analysis of the obtained phone embeddings, and show that cross-modal visual input can improve the discriminability of phonological features which are visually discernable (rounding, open/close, labial place of articulation), resulting in representations that are closer to abstract linguistic features than those based on audio only.
\end{abstract}


%Neil: pour les keywords retirerais aussi ABnet qui est redondant avec "Siamese network" et notre jargon à nous.

\section{Introduction}
The ability of many people, hearing and non-hearing, to lip read, demonstrates that speech perception is not only a purely auditory skill. Audio-visual integration is illustrated clearly by the McGurk effect \cite{McGurck}: the lip movements corresponding to [\textipa{g}], presented together with audio corresponding to a [\textipa{b}], is perceived as an intermediate sound, identified as [\textipa{d}], by many subjects. Such interactions between modalities have been documented in 5-month-old infants  \cite{Kuhl1984,McGurckInfants}. The visual channel for speech is poorer than the auditory channel, but provides information which can be complementary, especially regarding place of articulation \cite{Summerfield1992} (for example, between the coronal and labial consonants [\textipa{d}] and [\textipa{b}]).

Previous research has investigated the use of cross-modal information for speech recognition and has focused on systems trained with supervised learning (phoneme labels). Here, we investigate the case of weakly supervised learning, which is more appropriate for the modeling of infants' language acquisition. In particular, we will use a Siamese DNN architecture which feeds on word-level side information (the fact that two words are the same or different: \cite{Siamese,SynnaeveD14,zerospeech_abnet,kamper2016deep}), which demonstrably can be discovered automatically from continuous speech using spoken term discovery (\cite{zerospeech_abnet,levin2015segmental}).  In this paper, we use gold word-level annotations on the Blue-Lips corpus and focus on the combination of modalities in \emph{mono-task} and \emph{multi-task} settings.

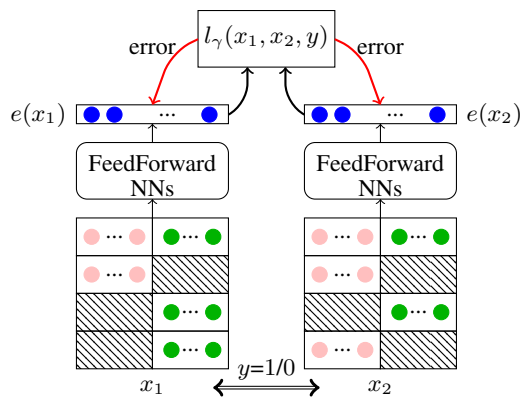
\begin{figure}
\centering
\caption{Multi-task setting. $x_1$ and $x_2$ are the training inputs and $y$ the output ($1$ if the inputs are the same words, and $0$ if not);  the pink circles are the acoustic features and the green circles the  visual ones. The shaded area are for zeroed inputs.\label{fig:Multitask}}
\begin{tikzpicture}
    \draw (9,22) rectangle (10,21.5);
    \draw[pink,fill=pink] (9.2,21.75) circle [radius = 0.1];
    \draw[pink,fill=pink] (9.8,21.75) circle [radius = 0.1];
    \node at (9.5,21.75) {...};    
    \draw (9,21.5) rectangle (10,21);
    \draw[pink,fill=pink] (9.2,21.25) circle [radius = 0.1];
    \draw[pink,fill=pink] (9.8,21.25) circle [radius = 0.1];
    \node at (9.5,21.25) {...};    
    \draw[pattern=north west lines, pattern color=black]
    (9,21) rectangle (10,20.5);
    \draw [pattern=north west lines, pattern color=black](9,20.5) rectangle (10,20);
    \draw (10,22) rectangle (11,21.5);
    \draw[codegreen,fill=codegreen] (10.25,21.75) circle [radius = 0.1];    
    \draw[codegreen,fill=codegreen] (10.8,21.75) circle [radius = 0.1];
    \node at (10.5,21.75) {...};    
    \draw[pattern=north west lines, pattern color=black]
    (10,21.5) rectangle (11,21);
    \draw (10,21) rectangle (11,20.5);
    \draw[codegreen,fill=codegreen] (10.25,20.75) circle [radius = 0.1];    
    \draw[codegreen,fill=codegreen] (10.8,20.75) circle [radius = 0.1];
    \node at (10.5,20.75) {...};    
    \draw(10,20.5) rectangle (11,20);
    \draw[codegreen,fill=codegreen] (10.25,20.25) circle [radius = 0.1];    
    \draw[codegreen,fill=codegreen] (10.8,20.25) circle [radius = 0.1];
    \node at (10.5,20.25) {...};

    \draw (12,22) rectangle (13,21.5);
    \draw[pink,fill=pink] (12.2,21.75) circle [radius = 0.1];
    \draw[pink,fill=pink] (12.8,21.75) circle [radius = 0.1];
    \node at (12.5,21.75) {...};    
    \draw (12,21.5) rectangle (13,21);
    \draw[pink,fill=pink] (12.2,21.25) circle [radius = 0.1];
    \draw[pink,fill=pink] (12.8,21.25) circle [radius = 0.1];
    \node at (12.5,21.25) {...};    
    \draw[pattern=north west lines, pattern color=black]
    (12,21) rectangle (13,20.5);
    \draw (12,20.5) rectangle (13,20);
    \draw[pink,fill=pink] (12.2,20.25) circle [radius = 0.1];
    \draw[pink,fill=pink] (12.8,20.25) circle [radius = 0.1];
    \node at (12.5,20.25) {...};    
    \draw (13,22) rectangle (14,21.5);
    \draw[codegreen,fill=codegreen] (13.25,21.75) circle [radius = 0.1];    
    \draw[codegreen,fill=codegreen] (13.8,21.75) circle [radius = 0.1];
    \node at (13.5,21.75) {...};    
    \draw[pattern=north west lines, pattern color=black]
    (13,21.5) rectangle (14,21);
    \draw (13,21) rectangle (14,20.5);
    \draw[codegreen,fill=codegreen] (13.25,20.75) circle [radius = 0.1];    
    \draw[codegreen,fill=codegreen] (13.8,20.75) circle [radius = 0.1];
    \node at (13.5,20.75) {...};    
    \draw[pattern=north west lines, pattern color=black]
    (13,20.5) rectangle (14,20);
    
    \node at (10,19.75) {$x_1$};\node at (13,19.75) {$x_2$};
    \draw[<->, double] (10.8,19.75) --(12.2,19.75);
    \node at (11.5,20){$y$=$1$/$0$};

    \draw[->] (10,22) --(10,22.25);
    \draw[rounded corners] (9,22.25) rectangle ++(2,0.75);
    \node at (10,22.73) {FeedForward};
    \node at (10,22.4) {NNs};
    
    \draw[->] (13,22) --(13,22.25);
    \draw[rounded corners] (12,22.25) rectangle ++(2,0.75);
    \node at (13,22.73) {FeedForward};
    \node at (13,22.4) {NNs};
    
    \draw[->] (10,23) --(10,23.25);
    \draw (9,23.25) rectangle (11,23.5);
    \draw[blue,fill=blue] (9.2,23.37) circle [radius = 0.1];
    \draw[blue,fill=blue] (9.5,23.37) circle [radius = 0.1]; \draw[blue,fill=blue] (10.75,23.37) circle [radius = 0.1];
    \node at (10.2,23.37) {...};    
    \node at (8.5,23.37) {$e(x_1)$};

    \draw[->] (13,23) --(13,23.25);
    \draw (12,23.25) rectangle (14,23.5);
    \draw[blue,fill=blue] (12.2,23.37) circle [radius = 0.1];
    \draw[blue,fill=blue] (12.5,23.37) circle [radius = 0.1]; \draw[blue,fill=blue] (13.75,23.37) circle [radius = 0.1];
    \node at (13.2,23.37) {...};    
    \node at (14.5,23.37) {$e(x_2)$};
      
    \draw [->, thick,rounded corners=8pt] (11,23.37)--(11.25,23.5)--(11.25,24);    
    \draw [->, thick,rounded corners=8pt] (12,23.37)--(11.75,23.5)--(11.75,24);
    \draw (10.6,24) rectangle (12.4,24.75);
    \node at (11.5,24.4) {$l_{\gamma}({ x_1,x_2, y})$};
    \draw [->, red, thick,rounded corners=8pt] (10.6,24.37)--(10.25,24.2)--(10,23.5);  
    \node at (10,24.25) {error};
    \draw [->, red, thick,rounded corners=8pt] (12.4,24.37)--(12.75,24.2)--(13,23.5);
    \node at (13,24.25) {error};
\end{tikzpicture}
\end{figure}

In mono-task settings, we train a Siamese network (see below) on only one type of input (audio only, visual only, or the concatenation of the two modalities). The multi-task setting uses a form of data augmentation in which we selectively knock out one of the modalities by setting all of the input features to zero. We include the four combinations shown in Figure \ref{fig:Multitask}, all of which are presented to the network during training, with equal probability. In the first three cases, the input to the two branches of the Siamese network is the same; the fourth case can be thought of as a lip-reading task, in which the network is presented with different modalities in each branch.

In  the  following sections, we first discuss related work. Then, we describe our model and the different evaluation measures. Finally, we report experimental results and conclude.

\section{Related work}
Classical audio-visual ASR systems use an audio-visual corpus to build complex supervised classifiers able to have an efficient phoneme representation \cite{duchnowski1994see,Noda2015}. While this approach benefits from a  rich phoneme representation that leads to good speech recognition, it needs thousands of hours of annotated speech, which is implausible as a model of language acquisition. Some other studies involve both geometry-based and appearance-based features to build less complex models \cite{Wu2016} but rely on an important upfront knowledge of optimal features. Other studies tackle the automatic lipreading task such as \textit{WLAS} \cite{Lipreading} and \textit{Lipnet} \cite{Lipreading2}. Those recent architectures yield state-of-the-art performance on lip reading. However, they focus primarily on the lip reading task (hence only on the acoustic contribution to the visual inputs), and are in a supervised setting. 

Our study deals with audio-visual ASR in a weakly supervised setting and evaluates the contribution of each modality in speech recognition. One of the most closely related works is \cite{Ngiam_multimodaldeep}, which learns an audio-visual speech representation in an unsupervised setting, but proposes to train a supervised classifier to evaluate the learned representation. As discussed in \cite{schatz_2016}, the supervised classification performance obtained on features is not a reliable indicator of the performance of unsupervised algorithms. Supervised learning may improve certain weaknesses in the features, such as poor scaling or noisy channels, which would present major obstacles in an unsupervised setting when using clustering algorithms. Rather, in our study, we evaluate different properties of the representation: its phonetic discriminability and its parallelism. Those measures will be introduced in the following sections.

Another relevant model is the DCCA \cite{DCCA, Wang2015} which can learn complex non-linear transformations of two data views to give a highly correlated embedding representation. However, maximizing the correlation in two views, in our case, the acoustic and visual views of the speech, would not necessarily lead to an efficient phoneme representation. Indeed, the acoustic signal can fully explain phonetic class identity, while it is under-specified in the visual modality. Hence, maximizing the correlation between these two modalities may lead to a loss of information for phonemes that share the same visual correlates.
%Neil: La dernière phrase on peut pas juste l'énoncer, faut la justifier. J'imagine que l'idée c'est qu'une modalité étant beaucoup moins fiable que l'autre on n'a pas intérêt à maximiser leur corrélation. Mais faut l'expliquer.

We use an ABnet type architecture \cite{SynnaeveD14}, a particular type of Siamese network that allows learning phonetic-level embeddings from word-level annotations. Such an architecture previously showed good performance \cite{zerospeech_abnet} in the context of the Zero Resource Speech Challenge 2015 \cite{zerospeech2015}.

\section{Methods}
\subsection{Dataset}
We use the Blue-Lips database \cite{BL}, an audio-visual speech corpus composed of 238 French sentences read by 16 speakers of Hexagonal French. Each speaker's recordings run around 20 minutes.

We represent the audio signal with 40 MFCCs computed from 40 mel-scale filterbanks sampled at 100 frames per second, resulting in a 40 dimensional vector. These 40 coefficients are the first 40 cepstral coefficients and do not contain delta/delta-delta nor energy features.

We use two kinds of visual features. First, we use dimensionality-reduced pixels from the region of the image corresponding to the mouth, identified using a Haar cascade classifier \cite{viola2001robust} to detect the mouth at the first frame and the meanshift algorithm \cite{MeanShift} to track the mouth throughout the rest of the video.
We match the video and audio frame rates by converting the video from 25 to 100 fps using ffmpeg.\footnote{https://ffmpeg.org/ffmpeg.html} This results in sequences of four identical frames for each original frame of video. The pixels for the mouth are spatially down-sampled to $30\times50$ pixels then whitened and reduced to 40 dimensions using PCA. We found that this did not reduce performance, consistent with \cite{duchnowski1994see}. Second, we concatenate these video features with lip landmarks, extracted using the active apperance model \cite{AAM}, a facial alignment algorithm which gives the shape of the mouth in 20 two-dimensional points. To match the audio frame rate here we apply cubic interpolation.

Finally, we apply mean-variance normalization to the features, for both modalities.

\subsection{Siamese network and ABnet}
A Siamese network is an architecture that contains two identical copies of the same network, so that the two subnetworks have the same configuration and share the same parameters.

In our experiments, we use ABnet, a particular Siamese network architecture. The model is represented in Figure \ref{fig:Multitask}. It  uses  pairs  of  words  to  learn  a  representation  of phones. The input to the network during training consists of pairs of stacked features $x_1$ and $x_2$ and a label $y \in \{0,1\}$, where $y=1$ if $x_1$ and $x_2$ represent the same word, and $y=0$ otherwise.  $x_1$ and $x_2$ represent stacks of frames that are in correspondence across the two words: for same-word pairs, this corresponence is the result of an alignment using dynamic time warping (DTW) \cite{dtw}, and, for different-word pairs, an alignment along the diagonal.

The idea of this architecture is that, given an abstract notion of similarity $D$, we can learn a representation in which the distance between the two embedding-space representations $e(x_1)$ and $e(x_2)$ reflects the similarity between the inputs $x_1$ and $x_2$: we want $D(e(x_1), e(x_2))$ to be small if $x_1$ and $x_2$ represent the same word, and large otherwise. In order to achieve this goal, the ABnet is trained with the margin cosine loss function:

\abovedisplayskip=0pt\relax
\[
l_{\gamma}({x_1,x_2, y})=
\begin{cases}
-cos(e(x_1), e(x_2)) & \text{if } y=1\\    max(0,cos(e(x_1), e(x_2))- \gamma) & \text{otherwise}\    
\end{cases}
\]
where $\gamma$ is the margin. 

In our experiments, we took a margin of $0.5$. Each of the inputs $x_1$ and $x_2$ consists of $7$ stacked frames of one of the modalities, or of the concatenation of the modalities. Each subnetwork contains $5$ hidden layers of $1000$ units with ReLU activations, and two output embeddings each of $39$ dimensions.

\subsection{Evaluation}

We evaluate two aspects of the speech representation: its ability to discriminate between phonemes, and its internal structure, which we probe to see whether phonological features are clearly coded.

\subsubsection{ABX task}
To evaluate the phonetic discriminability of our embeddings, we use an ABX discrimination task \cite{ABX, ABX2}. The task consists in presenting three stimuli $A$, $B$ and $X$, with $A$ and $B$ belonging to two different phonetic categories, and $X$ belonging to one of those categories (concretely, always $A$). Given a measure of divergence $D$ (not necessarily a proper distance metric), if $D(A,X)<D(B,X)$, then the score is $1$ (success), and otherwise the score is $0$ (failure).

In our experiments, $A$ and $B$ are minimal pairs of triphones: they are pairs of sounds composed each of three phonemes and differing only in their central phoneme (for example, \emph{beg,} /\textipa{bEg}/ and \emph{bag,} /\textipa{b\ae g}/, although the triphones need not be words). The first task is a within speaker task (WST), where the three stimuli are uttered by the same speaker. It measures how the relative distance between speech utterances in the embedding space correlates with their phonetic content. The second task is an across speaker task (AST), where $A$ and $B$ are uttered by the same speaker and $X$ by a different speaker. This task is harder than the previous one since it requires embeddings to mostly reflect the phonetic content despite the speaker change, and thus it requires invariance to speaker identity. In Table \ref{tab:abx}, we illustrate the tasks.

\begin{table}
    \centering
        \caption{Different ABX tasks, where $A$ and $X$ belong to the same category\label{tab:abx}}
        \begin{tabular}{rcccc}\hline
          Task & $A$ & $B$ & $X$ \\
          \hline
          WST & $/bag/_{talker1}$ & $/beg/_{talker1}$ & $/bag/_{talker1}$ \\
          AST & $/bag/_{talker1}$ & $/beg/_{talker1}$ & $/bag/_{talker2}$ \\

          \hline
\end{tabular}
\end{table}

A final score is obtained by averaging ABX scores over all the triplets in the corpus that were tested. In the overall ABX score below, we present the error rate ($1 - $ the ABX accuracy score), where an error of zero indicates a representation in which categories are perfectly separated, and an error of $50\%$ represents chance level. In the feature-by-feature ABX score, we use an ABX accuracy score to allow for comparability with the feature-by-feature parallelism analysis.

\subsubsection{Parallelism}

\begin{figure}
    \caption{A hypothetical two-dimensional representation displaying parallelism.\label{fig:Parallel}}
    \centering
        \adjincludegraphics[width=0.3\textwidth,trim={0 {.05\width} 0 {.06\width}},clip]{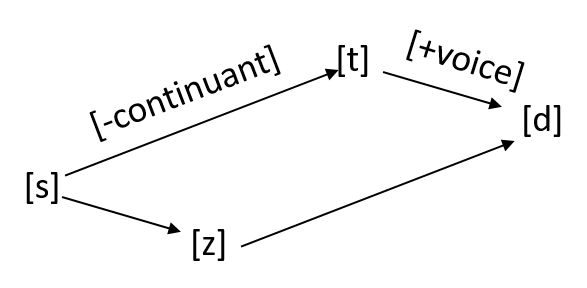}
\end{figure}

The second analysis measures how well the learned representations code individual phonological features, assessing the \emph{parallelism} of the representation \cite{Parallel}. For a given phonological feature, say [voice], representations of the feature are extracted by taking subtractions of phonemes that differ only in this feature (for [voice], [\textipa{d}] $-$ [\textipa{t}], [\textipa{z}] $-$ [\textipa{s}], and so on). In a space with perfect parallelism for voicing, these vectors will be exactly parallel (see Figure \ref{fig:Parallel}); the parallelism score we use here assesses relative parallelism, so that 
all subtraction vectors corresponding to a single feature need only be more parallel (have higher cosines) than pairs of subtraction vectors not corresponding to changes in the same single feature in order to obtain a maximal score (1). They will obtain a minimal score (0) when they are relatively more orthogonal or anti-collinear, and a score of 0.5 when they are no more parallel than vectors corresponding to different features. See \cite{Parallel} for details.

\section{Experiments}

\subsection{ABX discriminability}

Table \ref{tab:resultsABX} shows the ABX error rates within and across speaker for audio ($A$), visual ($V$) and the concatenated representation ($A$\&$V$). The distance used for the ABX tasks is the cosine distance. We notice first that, for all modalities, the ABnet embeddings have a lower error than the raw features. This was already demonstrated for the acoustic modality \cite{SynnaeveD14, Zeghidour2016, joint_learning}; we demonstrate that ABnet also improves the visual and concatenated input representations.

Tested on the visual modality, the ABnet improves the across-speaker ABX discriminability. In particular, the multi-task ABnet yields an embedding for visual information that is  more discriminative than the (raw) audio features, even though the audio signal itself is substantially more discriminative than the visual.  Furthermore, for the visual inputs, we reach the best performance with the multi-task model, demonstrating that this model takes advantage of the audio modality present in the training phase to learn a better visual representation.

For the audio modality, the mono-task model achieves the best performance. The presence of the visual modality during training deteriorates the embedding. 

\begin{table}[t]
\centering
\caption{Within- and across-speaker ABX discrimination \mbox{(\textbf{\%~error})} for audio, visual and multimodal representations by training condition (raw features, mono-task training, multi-task training).\label{tab:resultsABX}}
\setlength{\tabcolsep}{2.5pt}

\begin{tabular}{lccccccccc}
\toprule
            & \multicolumn{3}{c}{Raw features} &  & \multicolumn{3}{c}{Mono-task} &  &  \multicolumn{1}{c}{Multi-task} \\
 \cmidrule{2-4} \cmidrule{6-8} \cmidrule{10-10}\
Test        & A        & V        & A\&V       && A        & V       & A\&V       &&  A\&V,A,V,A\&V\\
\midrule
\multicolumn{10}{c}{Within speaker}\\
A           &  16.70   &    -     &      -     &&\bf{8.84} &   -     &    -       &&  9.80 \\
V           &   -      & 27.54    &     -      &&   -      &26.62    &  -         &&\bf{21.78}\\
A\&V        &   -      &    -     &     25.27  &&   -      &    -    & 10.76      &&\bf{10.53}\\
\midrule
\multicolumn{10}{c}{Across speaker}\\
A           &  26.12   &    -     &      -     &&\bf{11.44}&   -     &    -       && 12.39 \\
V           &   -      & 40.03    &     -      &&   -      &27.12    &  -         &&\bf{24.10}\\
A\&V        &   -      &    -     &     38.33  &&   -      &    -    &\bf{13.02}  && 14.01 \\
\bottomrule
\end{tabular}
\end{table}

To better understand the differences in the ABX scores, we examine the scores for particular phoneme pairs. We look at the pairs that correspond to  \emph{minimal oppositions} for phonological features: pairs of sounds that differ only in the given feature. Table \ref{phonefeature} lists the oppositions relevant to Hexagonal French. We separate the phonological features into \emph{visual} features---\textbf{Round} and \textbf{Coronal/Labial}---which correspond to features marked by lip rounding, and thus with clear visual correlates, and the remaining features, which we do not expect to have such strong visual correlates, and which we thus call \emph{non-visual} features.

\begin{table}[t]
\centering
\caption{Pairs of phonemes representing minimal oppositions for phonological features for French. \label{phonefeature}}
\setlength{\tabcolsep}{2.5pt}

\begin{tabular}{lp{4cm}}
  Phonological feature & Pairs \\
  \hline
  Round & [\textipa{e}]-[\textipa{\o}], [\textipa{E}]-[$\oe$], [\textipa{i}]-[\textipa{y}], [\textipa{\~A}]-[\textipa{\~O}]\\
  Coronal/Labial & [\textipa{d}]-[\textipa{b}], [\textipa{s}]-[\textipa{z}], [\textipa{t}]-[\textipa{p}], [\textipa{z}]-[\textipa{v}], [\textipa{n}]-[\textipa{m}]\\
  \hline
  Coronal/Dorsal & [\textipa{d}]-[\textipa{g}],[\textipa{t}]-[\textipa{k}]\\
  Continuant & [\textipa{b}]-[\textipa{v}],[\textipa{t}]-[\textipa{s}],[\textipa{d}]-[\textipa{z}],[\textipa{p}]-[\textipa{f}]\\
  Nasal vowel (NasalV)& [\textipa{a}]-[\textipa{\~A}], [\textipa{E}]-[\textipa{\~E}], [\textipa{O}],[\textipa{\~O}]\\
  Nasal consonant (NasalC)& [\textipa{b}]-[\textipa{m}],[\textipa{d}]-[\textipa{n}]\\
  Mid/High & [\textipa{e}]-[\textipa{i}],[\textipa{\o}]-[\textipa{y}],[\textipa{o}],[\textipa{u}]\\
  Voice & [\textipa{p}]-[\textipa{b}], [\textipa{t}]-[\textipa{d}], [\textipa{f}]-[\textipa{v}], [\textipa{k}]-[\textipa{g}], [\textipa{s}]-[\textipa{z}], [\textipa{S}]-[\textipa{Z}]\\
  \hline
\end{tabular}
\end{table}

\begin{figure}
    \caption{ABX accuracy, by phonological feature, for different embeddings.\label{fig:ABX_discrim}}
    \centering
    \adjincludegraphics[width=0.5\textwidth,trim={0.01 0 0 {0.05\width}},clip]{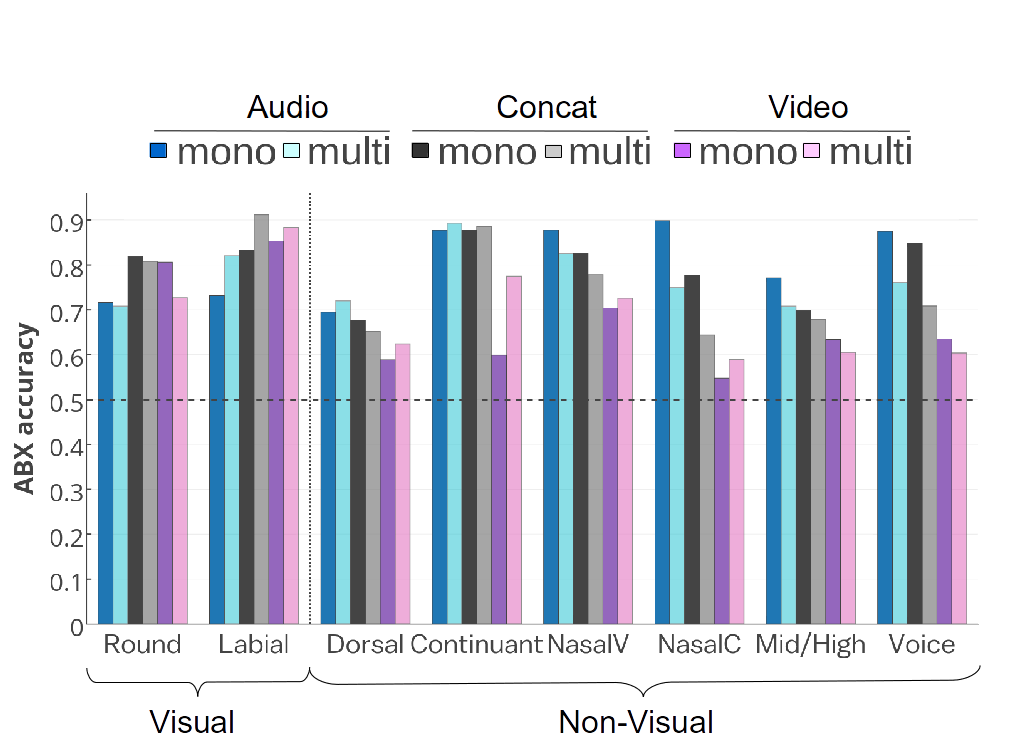}
\end{figure}

Figure \ref{fig:ABX_discrim} reports the ABX discriminability (in \textbf{\%~accuracy}) for the pairs in Table \ref{phonefeature}. We notice that for the non-visual phonological features, the mono-task audio nearly always has the highest ABX accuracy, except in two cases (\textbf{Dorsal} and \textbf{Continuant}) where the addition of visual information in training (multi-task audio) improves the contrasts slightly. In other words, in this case, representations that include visual information are handicapped with respect to phone discriminability for non-visual features. On the other hand, for the visual phonological features, the audio embeddings have the lowest ABX accuracy. The concatenated embeddings have the highest accuracy. Nevertheless, on average, the multi-task audio embedding gains discriminability compared to mono-task audio embedding for the consonant feature \textbf{Coronal/Labial}. Thus, for this feature, the audio inputs benefit from the visual modality present during the training. 

\subsection{Parallelism}

\begin{figure}
    \caption{Parallelism score, by phonological feature, for different embeddings.\label{fig:parallel_discrim}}
    \centering
  \adjincludegraphics[width=0.5\textwidth,trim={0 0 0 {.05\width}},clip]{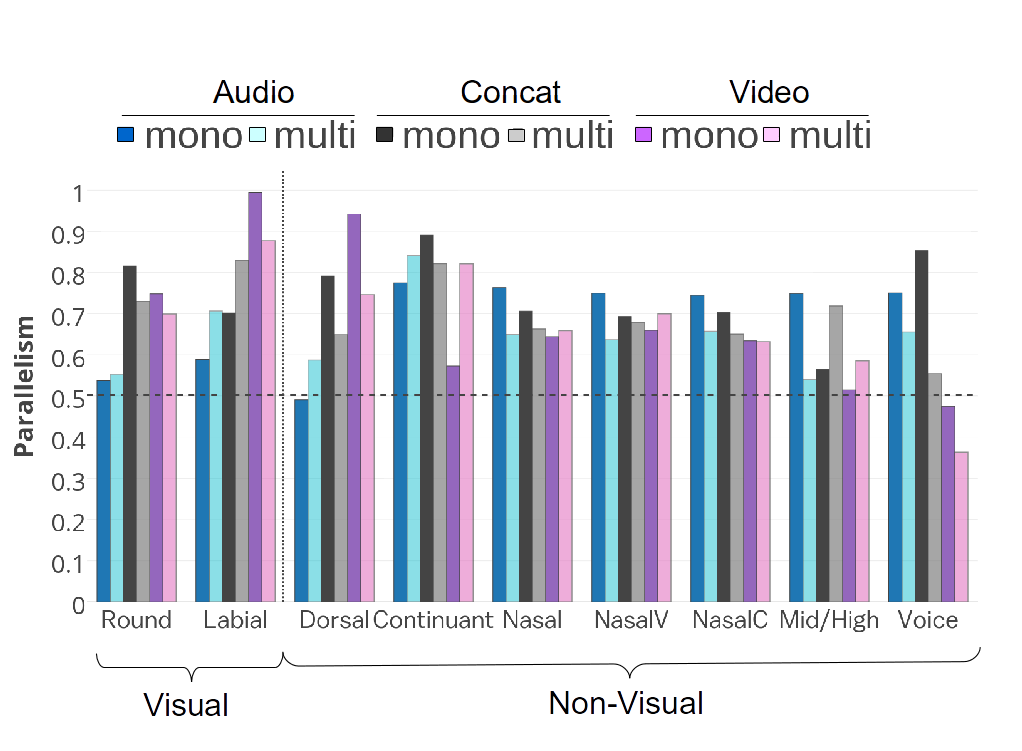}
\end{figure}

We now use the same set of minimal oppositions for the parallelism analysis discussed above, which examines the internal structure of the representations. Figure \ref{fig:parallel_discrim} shows the parallelism scores for the various embeddings. As previously, representations including visual information substantially improve the scores for visual features, indicating that these features have a more consistent representation in these embeddings. For non-visual features, the multi-task audio generally performs better than mono-task audio. However, unlike for the ABX analysis, an acoustic embedding has the best score only for three features (\textbf{Nasal vowel,} \textbf{Nasal consonant,} \textbf{Mid/high}). Thus, even for ``non-visual'' phonological features---those for which no obvious lip movement is expected---representations incorporating visual information have slightly more consistent encodings of these features.  On average, the mono-task concatenation embedding has the best parallelism, and is thus the best approximation of a phonological feature representation.

\subsection{McGurk effect}

To verify the plausibility of our models and to test if they can mimic the learning stages of the infant, we can see if they show the same patterns of audio-visual integration as human beings: we test whether presenting an audio signal corresponding to [\textipa{b}] with a mismatched visual signal, corresponding to [\textipa{g}], is perceived by the model as a [\textipa{d}] (the McGurk effect).

We perform an ABX discrimination task to see if these mismatched multi-modal inputs are unexpectedly similar to (hard to distinguish from) [\textipa{d}] (audio and video matched). To assess this, we construct three test sets of multi-modal inputs. The first matches the acoustic [\textipa{b}] with the same, generic visual held-out [\textipa{b}]. The second replaces this with a generic visual held-out [\textipa{p}]. The first gives a baseline score, and the second gives a combination which is mismatched, but which should not show integration effects (the lip movements are the same as for [\textipa{b}]). Finally, the last set is made to simulate the McGurk setting, replacing the visual [\textipa{b}] with a generic held-out [\textipa{g}]. We see that the ABX accuracy scores are lower in the McGurk case (Table \ref{tab:mcgurck}).

\begin{table}
    \centering
        \caption{ABX discriminability score between [\textipa{b}] and [\textipa{d}] \label{tab:mcgurck}}
        \begin{tabular}{lccc}\hline
          Task & Visual [\textipa{b}] & Visual [\textipa{p}] & Visual [\textipa{g}] \\
          \hline
          Mono-task concat & $77.93\%$ & $78.01\%$ & $65.47\%$ \\
          Multi-task concat & $93.00\%$ & $92.67\%$ & $79.60\%$ \\
        \hline
\end{tabular}
\end{table}
\section{Discussion}

This study introduces methods for learning multimodal speech representations in a weakly supervised setting. We use measures of the speech representations' performance on phone discrimination, and introduce analyses of the internal structure of the representations. The discriminability analysis shows that weakly supervised learning using ABnet improves phoneme discriminability over the input features in all cases, and that, for certain phonemic contrasts (those with strong visual correlates), adding visual information helps in discrimination. It also changes the structure of the representation to give more coherent representations of the relevant phonological features. The model can take advantage of the visual information even when it is present only during training. For phonological contrasts which do not benefit from visual information, the methods developed here for adding visual information reduce discriminability in some cases. This shows that visual information should only be exploited selectively, and discarded when it does not provide discriminative information. This suggests future research incorporating into our models a gating or attention system \cite{attentionmodel} that would learn when to use each modality, or both, in order to dynamically ignore uninformative features at test time.

\section{Acknowledgment}
This work was supported by the European Research Council (ERC-2011-AdG-295810 BOOTPHON), the Agence Nationale pour la Recherche (ANR-10-LABX-0087 IEC, ANR-10-IDEX-0001-02 PSL*), the Ecole de Neurosciences de Paris, the Region Ile de France DIM Cerveau et pensée, and an Amazon Web Services in Education Research Grant award.
\bibliographystyle{IEEEtran}

\bibliography{bibliography.bib}

\end{document}